\documentclass[10pt]{article}
\usepackage[letterpaper]{geometry}
\usepackage{hicss}
\usepackage{times}
\usepackage[none]{hyphenat}
\usepackage{url}
\usepackage{latexsym}
\usepackage{indentfirst}
\usepackage{graphicx}
\graphicspath{{images/}}
\usepackage{booktabs}
\usepackage[style=apa,]{biblatex}
  
\usepackage{tabularx}
\usepackage{amsmath}
\usepackage{amssymb}
\usepackage{amsfonts}
\usepackage{adjustbox}

\usepackage{mathtools, nccmath}
\usepackage{subcaption}
\usepackage[skip=5pt]{caption}
\usepackage{xpatch}
\xpatchcmd{\NCC@ignorepar}{%
\abovedisplayskip\abovedisplayshortskip}
{%
\abovedisplayskip\abovedisplayshortskip%
\belowdisplayskip\belowdisplayshortskip}
{}{}

\makeatletter
\newcommand{\showfloatsep}{
  \typeout{abovedisplayskip=\the\abovedisplayskip}
  \typeout{belowdisplayskip=\the\belowdisplayskip}
  \typeout{intextsep=\the\intextsep}
\typeout{Sectionspacingbefore=\the\@beforetitleskip}
\typeout{Sectionspacingafter=\the\@aftertitleskip}
}
\makeatother

\usepackage{todonotes}

\title{UWB-PostureGuard: A Privacy-Preserving RF Sensing System \\ for Continuous Ergonomic Sitting Posture Monitoring}

\author{
\begin{tabular}{cccc}
Haotang Li & Zhenyu Qi & Sen He & Kebin Peng\\
University of Arizona & University of Arizona & University of Arizona & East Carolina University\\
\underline{haotangl@arizona.edu} & \underline{qzydustin@arizona.edu} & \underline{senhe@arizona.edu} & \underline{pengk24@ecu.edu}
\end{tabular}
\\ \\
\begin{tabular}{ccccc}
Sheng Tan & Yili Ren & Tomas Cerny & Jiyue Zhao & Zi Wang\\
Trinity University & University of South Florida & University of Arizona & University of Georgia & Augusta University\\
\underline{stan@trinity.edu} & \underline{yiliren@usf.edu} & \underline{tcerny@arizona.edu} & \underline{jzhao@uga.edu} & \underline{zwang1@augusta.edu}
\end{tabular}
}

\date{}

\begin{document}
\maketitle

\begin{abstract}
Improper sitting posture during prolonged computer use has become a significant public health concern.
Traditional posture monitoring solutions face substantial barriers, including privacy concerns with camera-based systems and user discomfort with wearable sensors. This paper presents UWB-PostureGuard, a privacy-preserving ultra-wideband (UWB) sensing system that advances mobile technologies for preventive health management through continuous, contactless monitoring of ergonomic sitting posture. 
Our system leverages commercial UWB devices, utilizing comprehensive feature engineering to extract multiple ergonomic sitting posture features. 
We develop PoseGBDT to effectively capture temporal dependencies in posture patterns, addressing limitations of traditional frame-wise classification approaches. Extensive real-world evaluation across 10 participants and 19 distinct postures demonstrates exceptional performance, achieving 99.11\% accuracy while maintaining robustness against environmental variables such as clothing thickness, additional devices, and furniture configurations. 
Our system provides a scalable, privacy-preserving mobile health solution on existing platforms for proactive ergonomic management, improving quality of life at low costs. 
\end{abstract}

\subsubsection*{Keywords:}

Ultra-wideband sensing, posture monitoring, mobile health technology, privacy-preserving healthcare, preventive health management


\section{Introduction}\label{sec:introduction}

Poor sitting posture during prolonged computer use has become a critical ergonomic health concern, with profound implications for public health and economic burden. 
Recent studies reveal alarming statistics about posture-related health impacts: 70.5\% of university community members report musculoskeletal discomfort, particularly affecting the neck (86.4\%), lower back (75.9\%), and shoulders (76.2\%) \autocite{susilowati2022prevalence}, while 65.3\% of children and adolescents exhibit incorrect posture \autocite{li2020prevalence}. The health consequences extend beyond musculoskeletal issues, with poor posture promoting incontinence, constipation, and heartburn through increased abdominal pressure \autocite{harvard2023posture} and prolonged sitting significantly increasing mortality risk from heart disease and cancer \autocite{mayo2025sitting}. Work-related musculoskeletal disorders represent an economic burden equivalent to 2\% of the European Union's gross domestic product \autocite{garcia2024combining}.
This financial impact, combined with the widespread nature of postural problems, underscores the urgent need for effective, scalable monitoring solutions.

\begin{figure}
    \centering
    \includegraphics[width=\linewidth]{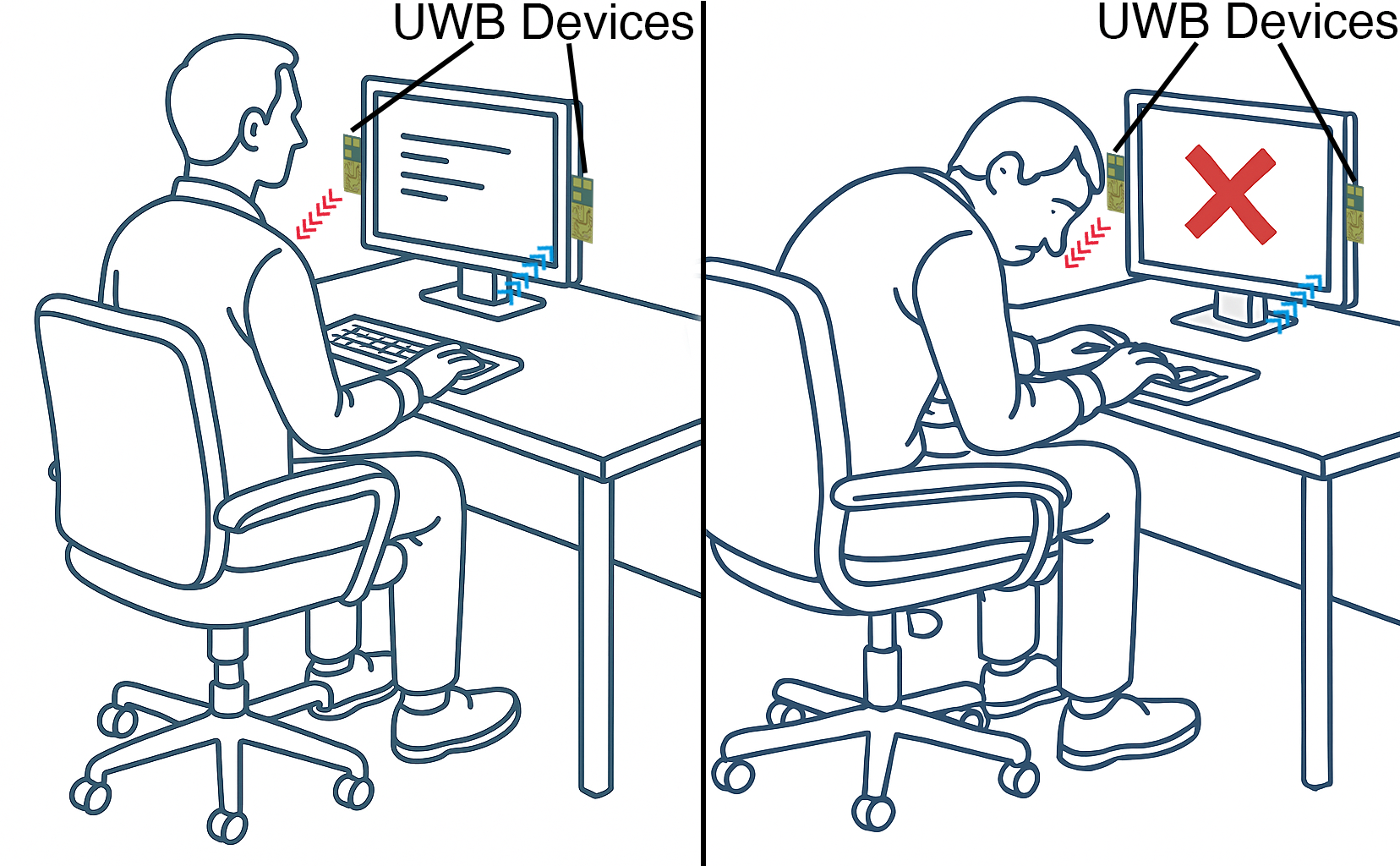}
    \caption{Continuous ergonomic sitting posture monitoring system.}
    \label{fig:core_idea}
\end{figure}
Traditional ergonomic sitting posture monitoring methods suffer from significant limitations that hinder their widespread adoption.
In particular, computer video-based approaches are often impractical due to privacy concerns and their dependence on specific sensor devices \autocite{wheelchair2024intelligent,sitting2024comprehensive}. Moreover, computer vision-based methods are also susceptible to variations in lighting, background conditions, and occlusions, which can lead to recognition errors \autocite{zhang2020survey}.
Wearable sensor-based approaches often cause user discomfort and require continuous usage, posing significant barriers to long-term adoption \autocite{smartsit2017sitting}. 

Radio frequency (RF) sensing offers a promising solution for posture monitoring due to its contactless nature, ability to operate in non-line-of-sight (NLoS) and low-light conditions, preservation of privacy, and ubiquitous network availability \autocite{wang2020human}.
Among RF signals, ultra-wideband (UWB) signals are particularly well-suited for medical sensing applications due to their high penetration capability, precise ranging, and low power consumption \autocite{rabbani2020overview}. 
Recent research shows UWB tag-based systems achieving high accuracy for activity recognition \autocite{automated2024detection}, while UWB positioning systems demonstrate centimeter-level accuracy for tracking dynamic activities \autocite{vandriel2018experimental}.
However, existing UWB-based posture research primarily focuses on basic activity recognition, with limited attention to detailed ergonomic analysis and insufficient validation in comprehensive workplace settings. This research gap and the urgent public health needs create a pressing demand for an approach that focuses on monitoring postural deviations from sitting and providing timely interventions that can deliver meaningful health benefits while maintaining the usability and privacy of the system.

This paper presents \textbf{UWB-PostureGuard} a privacy-preserving RF sensing system for real-time ergonomic health monitoring that advances beyond existing UWB applications through comprehensive feature engineering and robust temporal modeling. The key contributions include:
\textbf{1. Comprehensive UWB Feature Engineering:} We develop a novel feature extraction framework that leverages multiple dimensions of UWB data, including ranging measurements, Channel Impulse Response (CIR) magnitude and phase components, Angle of Arrival (AoA) estimations, and signal quality indicators, providing richer posture discrimination capabilities than previous single-metric approaches.
\textbf{2. Temporal-Aware GBDT Classification:} We design the PoseGBDT model that is enhanced with sliding window strategies, time-lagged features, and rolling statistical summaries to capture temporal dependencies in posture patterns, addressing the limitations of the standard frame-wise classification approaches.
\textbf{3. Comprehensive Real-World Evaluation:} We conduct extensive validation in real-world workplace environments with complex furniture configurations and environmental conditions, demonstrating system robustness and practical deployability beyond controlled laboratory settings.
\textbf{4. Privacy-by-Design Implementation:} We ensure complete user privacy through RF-only sensing without capturing identifiable personal information, addressing critical workplace surveillance concerns while maintaining high classification accuracy.

The remainder of this paper is organized as follows: Section~\ref{sec:related_work} reviews related work in posture monitoring and UWB sensing. Section~\ref{sec:preliminary} provides preliminary background on UWB technology and ergonomic posture assessment. Section~\ref{sec:system_design} details the system design, including scenario setup, feature engineering, and model design. Section~\ref{sec:exp_setup} elaborates on the data collection process and experiment setups.
Section~\ref{sec:evaluation} presents comprehensive evaluation results across multiple metrics and scenarios. Section~\ref{sec:discussion} discusses implications, limitations, and future directions. Section~\ref{sec:conclusion} concludes with key findings and contributions.

\section{Related Work}
\label{sec:related_work}
Recent advancements in sitting pose recognition have contributed significantly to the development of health monitoring systems, especially for elderly care and rehabilitation applications \autocite{sittingpose}. Traditional vision-based approaches, leveraging deep learning techniques such as convolutional neural networks, have achieved remarkable accuracy in sitting posture recognition; however, these methods often encounter challenges related to privacy concerns and varying lighting conditions \autocite{sittingpose_review}. 

To address these limitations, sensing technologies based on UWB devices have emerged as promising alternatives for healthcare environments due to their capability to penetrate obstacles and preserve user privacy \autocite{uwb_act_recog}. Recent studies have demonstrated that UWB sensing can robustly estimate human body positions and movements, including subtle posture changes, without requiring wearable devices \autocite{uwb_state_est}. Integrating machine learning with UWB data further enhances the accuracy of activity classification and pose estimation, thereby enabling continuous and contactless health monitoring \autocite{uwb_blood_glucose,uwb_blood_preasure}. Despite notable progress, challenges remain in continuous posture monitoring, and minimizing the system’s false detection rates, which continue to be active research directions in this field.

\section{Preliminary}
\label{sec:preliminary}
\textbf{UWB Sensing.} UWB device transmits extremely short pulses across a wide frequency spectrum (typically exceeding 500 MHz bandwidth), enabling sub-centimeter accuracy in distance measurement and fine-grained temporal resolution. These properties make UWB particularly suitable for detecting subtle posture changes and small-scale human movements. 

\textbf{Privacy-Preserving.} Compared to camera-based systems, UWB sensing offers significant privacy advantages by capturing only reflected electromagnetic signals without generating identifiable visual data. As a non-invasive technology, the sensitive biometric features, faces, and personal environments are never recorded or processed, greatly reducing privacy risks. Therefore, UWB signals are difficult to use for reconstructing detailed human shapes or actions.

\textbf{Channel Impulse Response.}
Its ability to penetrate common materials and maintain robustness against environmental noise ensures reliable performance in diverse indoor environments. In this study, the CIR data obtained from UWB signals play a key role in posture classification. CIR characterizes the multipath propagation of UWB pulses, capturing how the signal reflects off the human body and the surrounding environment. By analyzing CIR patterns, we found that different postures, such as upright sitting and hunched sitting, produce distinguishable CIR signatures. 
\begin{figure}[ht]
    \centering
    \begin{minipage}[t]{0.71\linewidth}
        \begin{subfigure}[t]{\linewidth}
            \includegraphics[width=\linewidth]{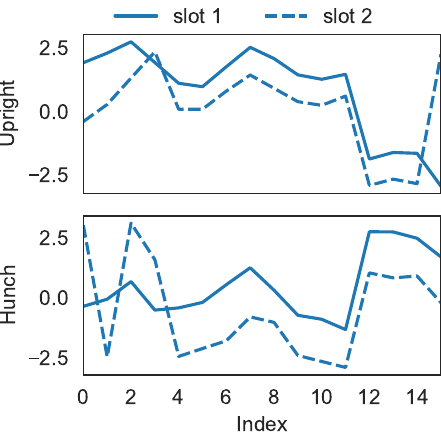}
        \end{subfigure}
    \end{minipage}
    \hfill
    \begin{minipage}[b]{0.27\linewidth}
        \begin{subfigure}[t]{\linewidth}
            \includegraphics[width=\linewidth]{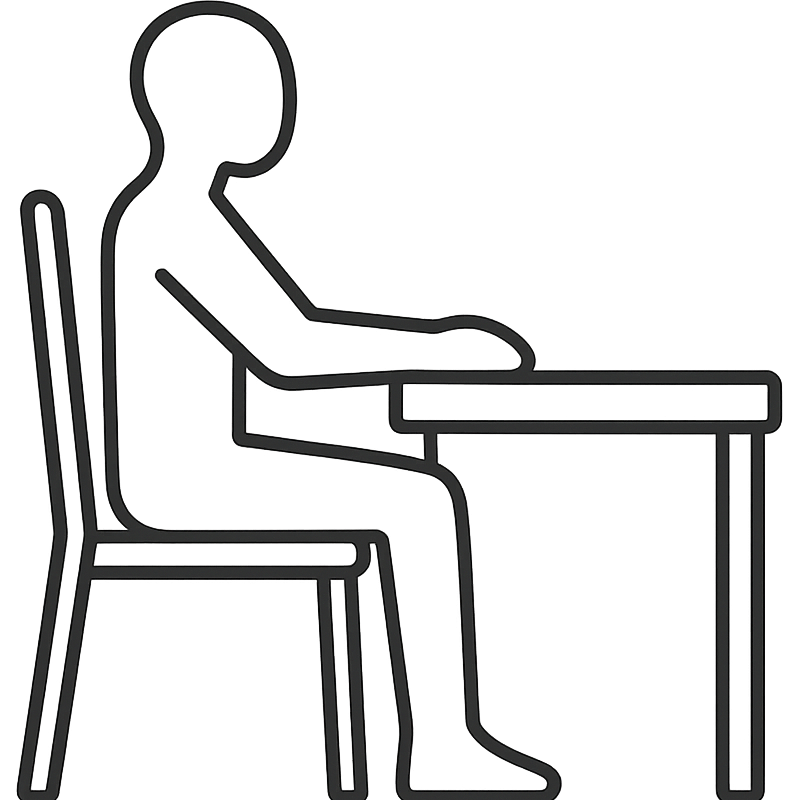}
        \end{subfigure}
        \vspace{7pt}
        \begin{subfigure}[t]{\linewidth}
            \includegraphics[width=\linewidth]{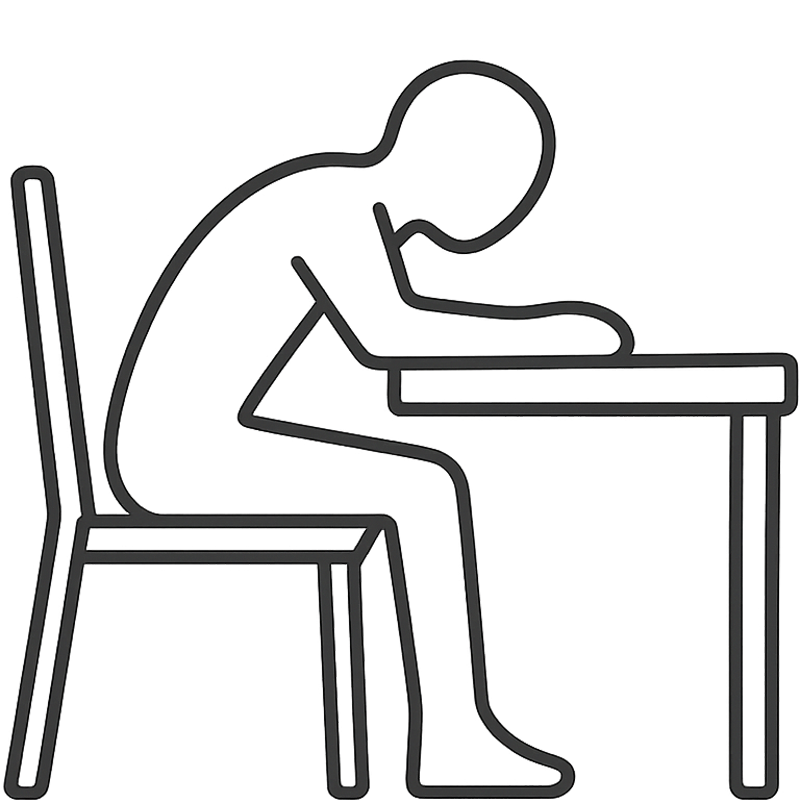}
        \end{subfigure}
        \vspace{5pt}
    \end{minipage}
    \caption{CIR comparison between different postures.}
    \label{fig:preliminary}
\end{figure}

Figure~\ref{fig:preliminary} shows the magnitude of CIR data from two time slots under two postures: upright sitting and hunching. For the same posture, different samples display similar CIR patterns, indicating strong intra-class consistency. In contrast, the patterns between upright and hunched postures are clearly different, reflecting distinct multipath propagation effects for each posture. This demonstrates that CIR captures posture-dependent yet stable features, making it well-suited for robust and privacy-preserving posture classification.

\section{System Design}
\label{sec:system_design}

\subsection{System Overview}
\label{subsec:sys_overview}

\begin{figure}[ht]
    \centering
    \includegraphics[width=\linewidth]{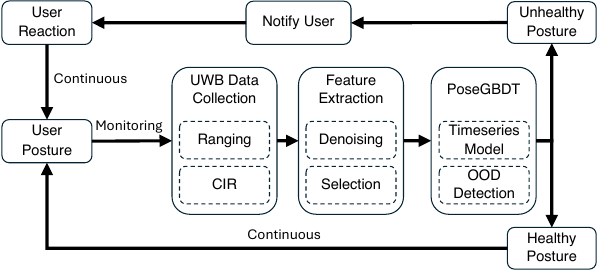}
    \caption{Overall system diagram.}
    \label{fig:sys-overview}
\end{figure}

Proposed UWB-PostureGuard system leverages commercial UWB devices to continuously monitor user posture and provide real-time feedback. As shown in Figure \ref{fig:sys-overview}, the system operates through a four-stage pipeline: (1) \textit{UWB Data Collection} captures ranging measurements and Channel Impulse Response (CIR) data from the user's sitting environment; (2)~\textit{Feature Extraction} processes the raw data through denoising and selection procedures to extract relevant postural features; (3)~The system then utilizes our temporal-aware \textit{PoseGBDT} model with integrated timeseries analysis and out-of-distribution (OOD) detection to distinguish between healthy and unhealthy postures; and (4) then the system alerts users when problematic postures are detected. The system maintains continuous monitoring throughout this process, creating a closed feedback loop where user reactions to notifications are incorporated back into the posture assessment, enabling adaptive and personalized ergonomic sitting health monitoring.

\subsection{Feature Engineering}
\label{subsec:feat_eng}

Our feature engineering framework leverages multiple UWB data streams to create a comprehensive postural signature that surpasses single-metric approaches used in prior work. We extract and integrate three primary data categories: ranging features, signal quality features, and Channel Impulse Response (CIR) based features.

We first collect \textit{ranging features} using Double-Sided Two-Way Ranging (DS-TWR), a time-of-flight protocol that provides distance, azimuth, and elevation angle measurements along with reliability metrics (Table \ref{tab:ranging_data}). Our system employs dual antenna pairs configured for orthogonal angle-of-arrival estimation, enabling spatial characterization of postural changes.

\begin{table}[ht]
    \scriptsize
    \centering
    \caption{Key ranging data attributes for posture classification.}
    \label{tab:ranging_data}
    \begin{tabularx}{\linewidth}{|p{0.12\linewidth}|X|}
        \hline
        \textbf{Attribute} & \textbf{Explanation} \\
        \hline
        Distance & Distance between Initiator and Anchor in cm. \\
        \hline
        Azimuth & Azimuth angle of arrival.   \\
        \hline
        Elevation & Elevation angle of arrival. \\
        \hline
        FOM & Reliability of estimated AoA, value from 0 to 100. Higher values indicate better estimate quality. \\
        \hline
        PDoA & Estimation of phase difference in degrees from antenna pair. \\
        \hline
    \end{tabularx}
\end{table}

\begin{table}[ht]
    \scriptsize
    \centering
    \caption{Explanation of attributes in antenna frames.}
    \label{tab:rframe_data}
    \begin{tabularx}{\linewidth}{|p{0.27\linewidth}|X|}
        \hline
        \textbf{Attribute} & \textbf{Explanation} \\
        \hline
        NLoS & Indicates if the ranging frame was in Line of sight or Non Line of Sight. \\
        \hline
        First Path Index & Estimated first path index (ns). \\
        \hline
        Main Path Index & Estimated main path index (ns). \\
        \hline
        SNR Main Path & Signal-to-Noise Ratio (SNR) of the main path (dB). \\
        \hline
        SNR First Path & SNR of the first path (dB). \\
        \hline
        SNR Total & SNR of configured RX (dB). \\
        \hline
        RSSI & Received Signal Strength Indicator (dB). \\
        \hline
        CIR Main Power & unsigned 32-bit value. \\
        \hline
        CIR First Path Power &  unsigned 32-bit value. \\
        \hline
        Noise Variance & Noise Variance in the CIR. \\
        \hline
        CFO & Carrier Frequency Offset estimate in PPM. \\
        \hline
        AoA Phase & Phase in degrees for calculating phase difference. \\
        \hline
    \end{tabularx}
\end{table}

Beyond basic ranging features, we extract detailed \textit{signal quality features} including SNR values, RSSI measurements, and path indices that capture environmental reflections caused by postural changes (Table \ref{tab:rframe_data}). This multi-dimensional approach provides richer discrimination capabilities compared to distance-only methods. In addition, our key contribution lies in exploiting Channel Impulse Response data to capture \textit{CIR-Based features} for fine-grained posture detection. The CIR data is processed as:

\begin{equation}
    \label{eq:cir}
    h[n] = \mathrm{Re}(h[n]) + j \cdot \mathrm{Im}(h[n]),
\end{equation}
where the real part $\mathrm{Re}(h[n])$ represents the in-phase component in the channel, and the imaginary part $\mathrm{Im}(h[n])$ represents the quadrature component. Having the CIR data in complex format, we calculate the magnitude value $|h[n]|$, which reflects the energy of each multi-path component; and the phase angle $\angle h[n]$, which captures the phase shift introduced by each propagation path.

We also implement Inter-quartile Range (IQR) outlier removal while preserving the inherent CIR denoising from the device's physical layer, balancing noise reduction with information preservation for optimal classification performance.

\begin{figure*}[ht]
    \centering
    \begin{subfigure}[t]{0.16\textwidth}
        \includegraphics[width=0.9\linewidth]{images/upright.png}
        \caption{Upright}
    \end{subfigure}
    \hfill
    \begin{subfigure}[t]{0.16\textwidth}
        \includegraphics[width=0.9\linewidth]{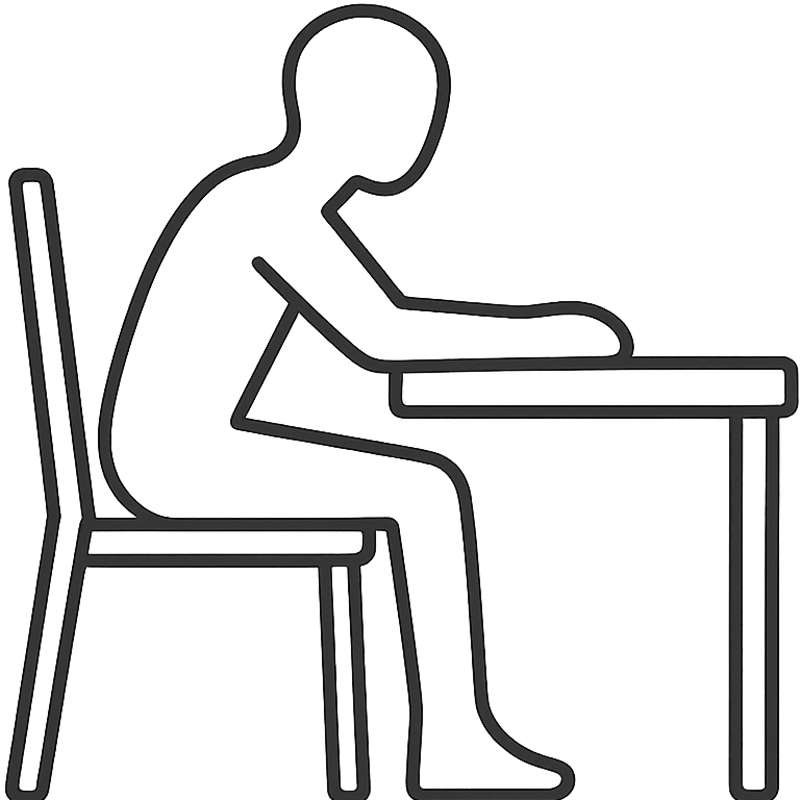}
        \caption{Lean Forward}
    \end{subfigure}
    \hfill
    \begin{subfigure}[t]{0.16\textwidth}
        \includegraphics[width=0.9\linewidth]{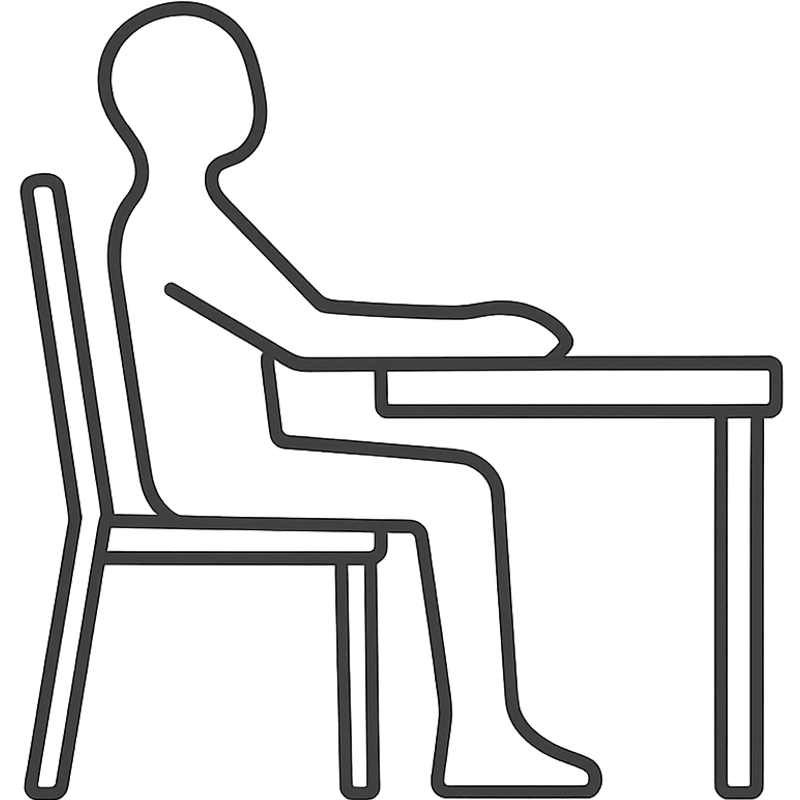}
        \caption{Lean Back}
    \end{subfigure}
    \hfill
    \begin{subfigure}[t]{0.16\textwidth}
        \includegraphics[width=0.9\linewidth]{images/hunch.png}
        \caption{Hunched}
    \end{subfigure}
    \hfill
    \begin{subfigure}[t]{0.16\textwidth}
        \includegraphics[width=0.9\linewidth]{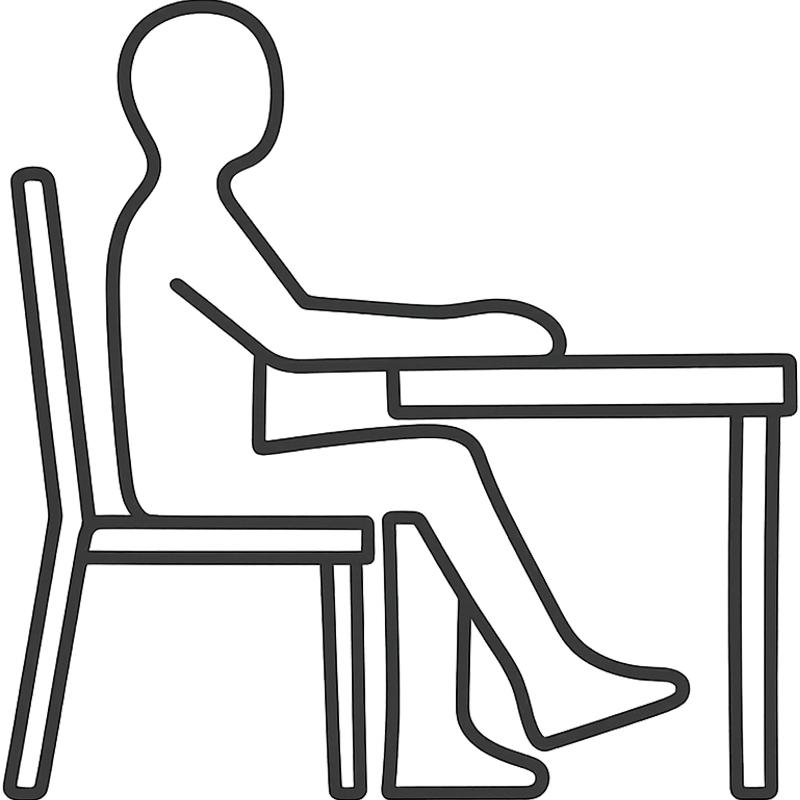}
        \caption{Cross Leg}
    \end{subfigure}
    \hfill
    \begin{subfigure}[t]{0.16\textwidth}
        \includegraphics[width=0.9\linewidth]{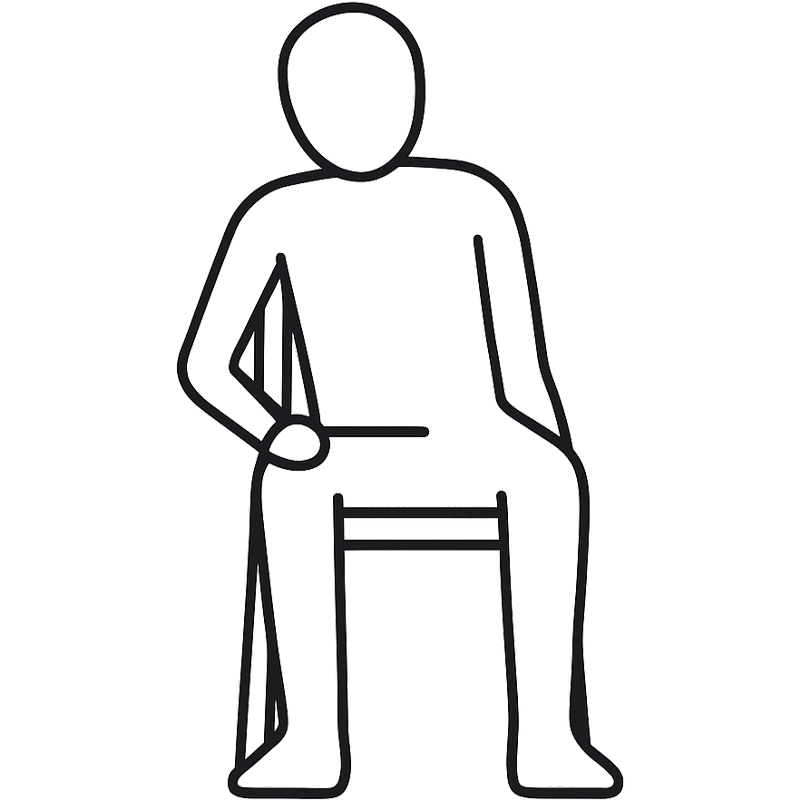}
        \caption{Lateral Lean}
    \end{subfigure}

    \medskip
    
    \begin{subfigure}[t]{0.16\textwidth}
        \includegraphics[width=0.9\linewidth]{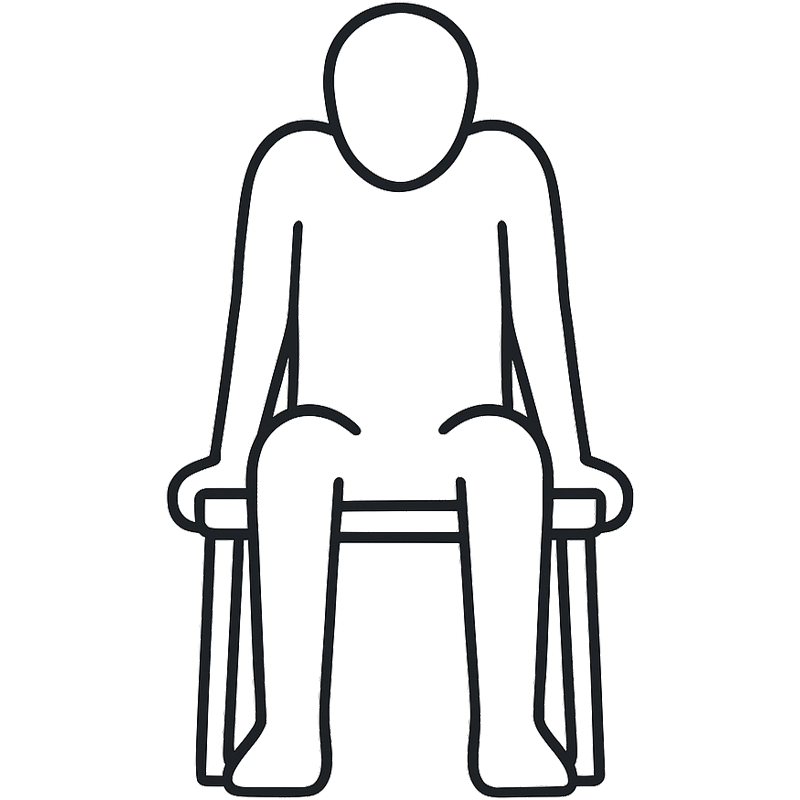}
        \caption{Tense}
    \end{subfigure}
    \hfill
    \begin{subfigure}[t]{0.16\textwidth}
        \includegraphics[width=0.9\linewidth]{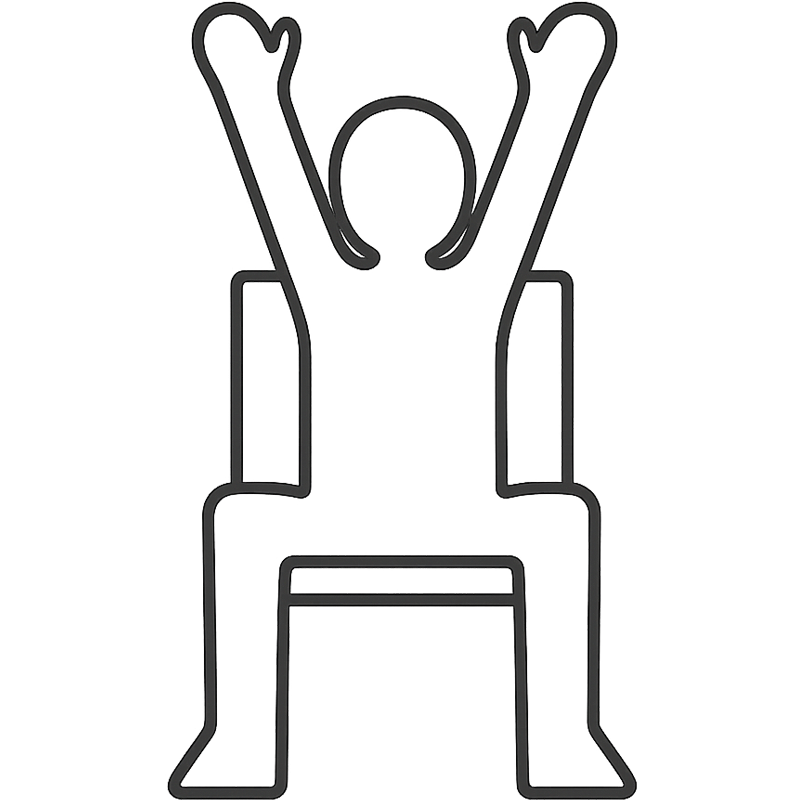}
        \caption{Stretch}
    \end{subfigure}
    \hfill
    \begin{subfigure}[t]{0.16\textwidth}
        \includegraphics[width=0.9\linewidth]{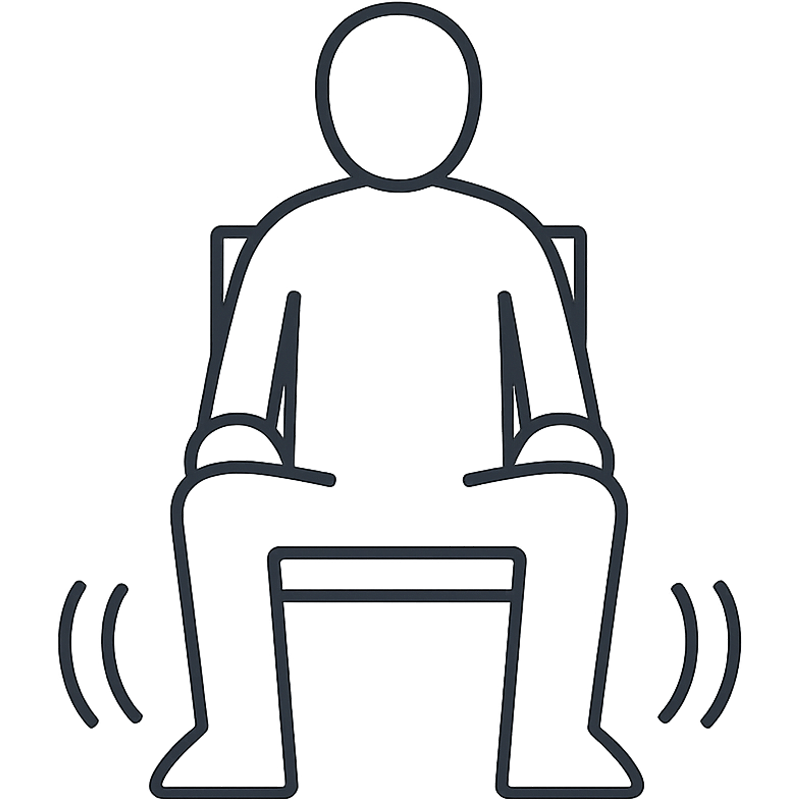}
        \caption{Horiz. Leg Shake}
    \end{subfigure}
    \hfill
    \begin{subfigure}[t]{0.16\textwidth}
        \includegraphics[width=0.9\linewidth]{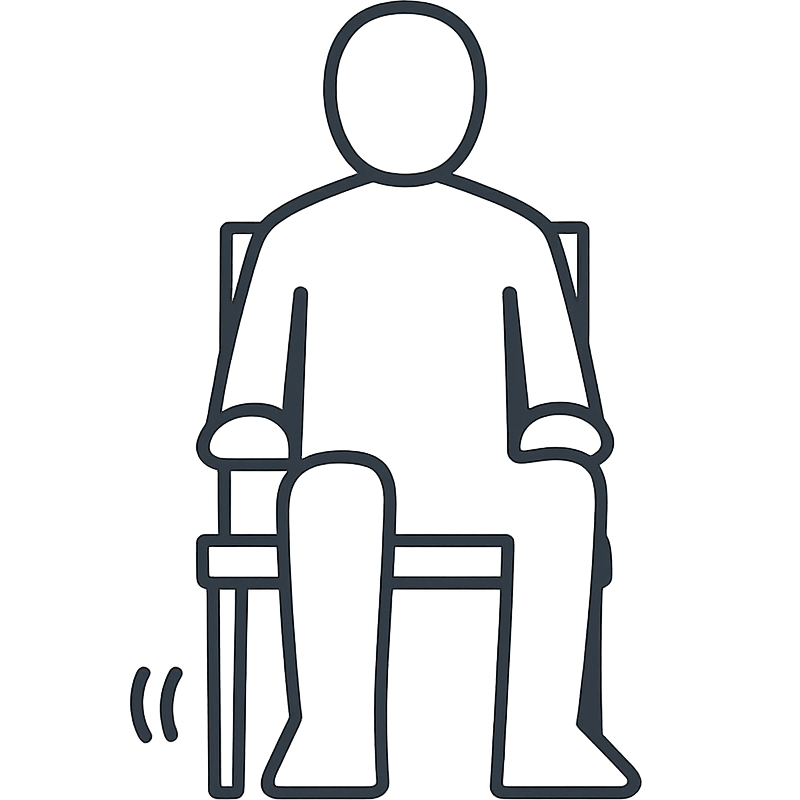}
        \caption{Vert. Leg Shake}
    \end{subfigure}
    \hfill
    \begin{subfigure}[t]{0.16\textwidth}
        \includegraphics[width=0.9\linewidth]{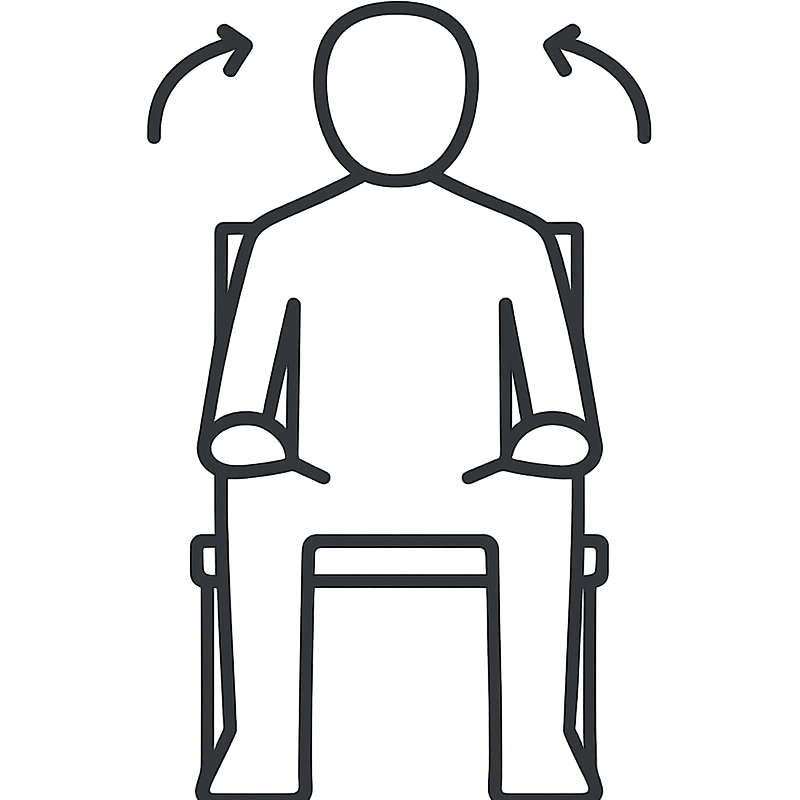}
        \caption{Rotate Head}
    \end{subfigure}
    \hfill
    \begin{subfigure}[t]{0.16\textwidth}
        \includegraphics[width=0.9\linewidth]{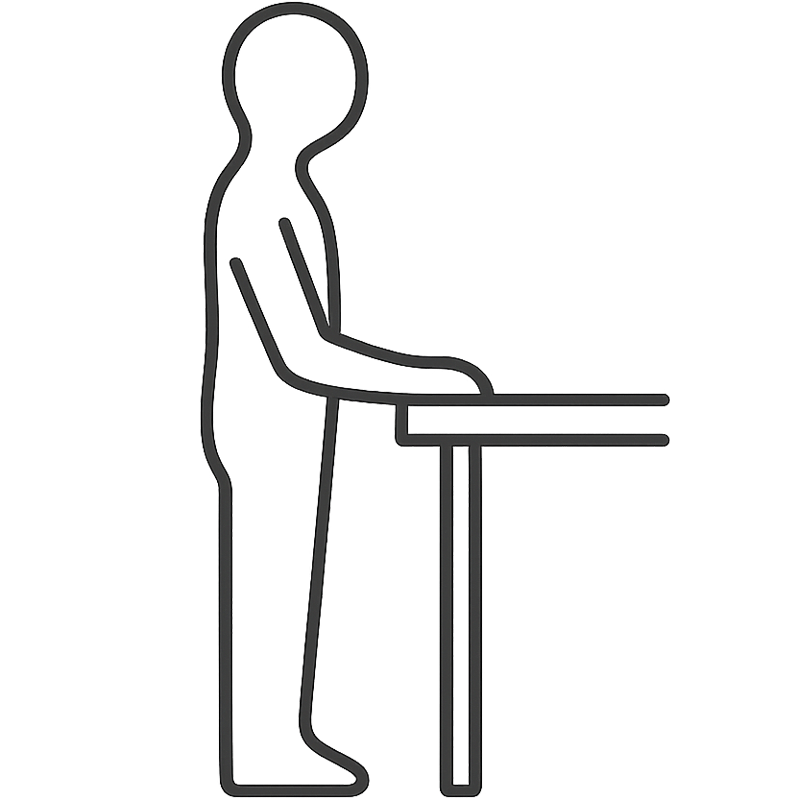}
        \caption{Stand}
    \end{subfigure}
    \caption{Partial examples of everyday sitting postures.}
    \label{fig:postures}
\end{figure*}

\subsection{Sitting Posture Taxonomy}
\label{subsec:posture}
The selected postures cover a wide range of everyday sitting behaviors commonly observed in office and healthcare settings, partially shown in Figure \ref{fig:postures}, aligning with ergonomic assessment tools such as the New York Posture Rating Chart \autocite{nyprc}. 
We include neutral postures (e.g., upright, cross leg (left/right)) to represent baseline sitting behavior, and intentional variations (e.g., leaning forward, leaning back, lateral lean) that reflect frequent posture shifts during prolonged sitting. We also incorporate more pronounced deviations such as hunched, tense, and lying on table, which may signal fatigue or improper ergonomics. To capture postural asymmetries and body dynamics, we include rotating the neck, and various forms of leg shaking and finger tapping, which are indicative of discomfort, stress, or restlessness. Additionally, stretch, stand, and walk transitions are added to capture activity boundaries and support broader behavior modeling. 
Notably, idle posture means no human is present before the devices.
This diverse set of 19 postures ensures that our dataset comprehensively reflects realistic and clinically meaningful variations in sitting behavior, providing a robust foundation for validating our UWB-based posture classification system.

\subsection{PoseGBDT Model}
\label{subsec:models}

\textbf{Model Architecture Design.} 
Our PoseGBDT model addresses two key challenges in UWB-based posture recognition: handling high-dimensional, heterogeneous feature spaces and capturing temporal dependencies in postural transitions. We select Gradient Boosting Decision Trees (GBDT) as our backbone due to their proven effectiveness with mixed-type features, robustness to noise, and ability to capture complex features relevant for UWB sensing data that combines ranging measurements, signal quality indicators, and CIR patterns.
To validate this design choice, we compare five representative methods across different algorithmic paradigms (Table \ref{tab:method_comparison}). LightGBM outperforms alternatives, achieving 0.81 F1-score compared to 0.67 for Random Forest and 0.27 for k-NN, confirming GBDT's suitability for our multi-modal UWB feature space.

\begin{table}[ht]
\centering
\caption{Performance comparison across methods.}
\label{tab:method_comparison}
\scriptsize
\begin{tabularx}{\linewidth}{l *{1}{>{\centering\arraybackslash}X} *{1}{>{\centering\arraybackslash}X} *{1}{>{\centering\arraybackslash}X} *{1}{>{\centering\arraybackslash}X}}
\toprule
\textbf{Posture} & \textbf{Precision} & \textbf{Recall} & \textbf{F1-Score} & \textbf{Accuracy} \\
\midrule
Na\"ive Bayes & 0.21 & 0.12 & 0.09 & 0.12 \\
$ k $-NN      & 0.29 & 0.92 & 0.27 & 0.28 \\
Random Forest & 0.67 & 0.68 & 0.67 & 0.68 \\
MLP           & 0.20 & 0.14 & 0.10 & 0.14 \\
\textbf{LightGBM}      & \textbf{0.81} & \textbf{0.81} & \textbf{0.81} & \textbf{0.81} \\
\bottomrule
\end{tabularx}
\end{table}

LightGBM outperforms all other methods in our system; thus, we select LightGBM as our backbone classification model. LightGBM is a Gradient Boosting Decision Tree (GBDT) method, which fits a decision tree at the $ m $-th step $ h_{m}(x) $ to pseudo-residuals. The tree partitions the input space into $ J_{m} $ disjoint regions $ R_{1m},\ldots ,R_{J_{m}m} $ and predicts a constant value in each region. The parameter update process of the GBDT model can be written as:
\useshortskip
\begin{equation}
    \label{eq:gbdt}
    F_m(x) = F_{m-1}(x) + \sum_{j=1}^{J_{m}} \gamma_{jm} \mathbf{1} _{R_{jm}}(x), 
\end{equation}
where $ J_{m} $ is the number of its leaves, $ b_{jm} $ is the value predicted in the region $ R_{jm} $, 
\useshortskip
\begin{align*}
    \gamma_{jm} = \operatorname*{arg\,min}_{\gamma} &\sum_{x_i \in R_{jm}} L(y_{i},F_{m-1}(x_{i}) + \gamma), \\ 
    h_{m}(x) = &\sum_{j=1}^{J_{m}}b_{jm} \mathbf{1}_{R_{jm}}(x) .
\end{align*}

\textbf{Temporal-Aware Enhancement.} 
The core innovation of our PoseGBDT lies in extending GBDT's inherent strength in feature interaction modeling to capture temporal patterns in postural behavior. Traditional posture classification approaches often lack leverage on the sequential nature of human postures, missing critical transition dynamics and contextual information. We address this through a comprehensive temporal modeling strategy.
Our sliding window approach enriches each frame with historical context using time-lagged features and rolling statistics:
\useshortskip
\begin{equation}
    \mathbf{x}_i^{(\text{shift}, k)} = \mathbf{x}_{i - k}, \quad \bar{\mathbf{x}}_i^{(w)} = \frac{1}{w} \sum_{j=0}^{w-1} \mathbf{x}_{i-j} \label{eq:shifting_rolling}
\end{equation}

This allows the model to implicitly learn temporal dependencies and improves its capacity to distinguish between time-dependent states.

\begin{figure}[ht]
    \centering
    \includegraphics[width=\linewidth]{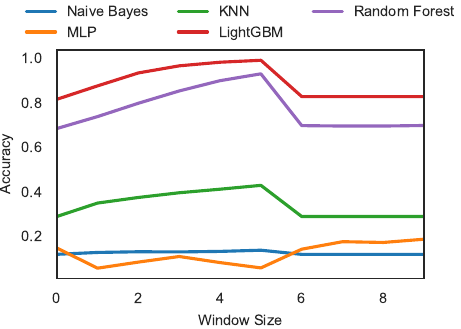}
    \caption{Comparison of different sliding window sizes.}
    \label{fig:ablation_swin}
\end{figure}

We set the sliding window size $ \tau $, which affects the amount of historical data included in each input data. 
Figure \ref{fig:ablation_swin} demonstrates the critical importance of temporal window sizing. Tree-based methods achieve optimal performance at $ \tau = 5 $, balancing historical context with computational efficiency. Beyond $ \tau = 7 $, performance degrades due to feature correlation and limited tree splitting capacity, validating our design choices.

Besides, KNN is sensitive to high dimensions and the curse of dimensionality causes no difference when high $ \tau $.
Na\"ive Bayes is very sensitive to feature independence, and thus a large $ \tau $ causes more noise.
While the figure shows the MLP might have better performance when $ \tau $ increases, this also leads to a significant increase in computational consumption.

\textbf{Robust Deployment with OOD Detection.}
We enhance the robustness of our system by incorporating an out-of-distribution (OOD) detection mechanism alongside the LightGBM classifier. While LightGBM excels at classifying known postures, it lacks native support for identifying inputs outside the trained categories—an essential capability for real-world deployment. To address this, we extract internal leaf embeddings from the trained LightGBM model, representing each sample’s traversal across the decision trees. 
Our integrated One-Class SVM mechanism extracts internal leaf embeddings from the trained LightGBM model, learning the manifold of known posture patterns. This enables robust detection of novel postures, transition states, and noisy inputs while maintaining high classification accuracy on trained categories.

By combining a powerful GBDT-based classification model, a time-aware feature engineering strategy, and an out-of-distribution detection mechanism, our PoseGBDT model enables the system to effectively capture temporal patterns in UWB data while ensuring reliable performance in real-world applications.

\section{Experimental Setup}
\label{sec:exp_setup}

\subsection{Scenario Setup}
\label{subsec:secnario_setup}
To evaluate UWB-PostureGuard, we set up a typical daily scenario involving study or office work. The environment is furnished with standard office items, including desks, chairs, a computer monitor, a keyboard, and a mouse. 
To challenge the system's robustness, we introduce variability by manipulating the presence of common objects, including a pair of earbuds, a smartphone (in-pocket/on-desk), a smartwatch, a pillow, a backpack, clothing, a laptop, a water bottle, two types of chairs; and four different UWB device layouts. 
Additional variables that may be encountered during real-world deployment are detailed in Section \ref{sec:ablation}.

\begin{figure}[ht]
    \centering
    \includegraphics[width=\linewidth]{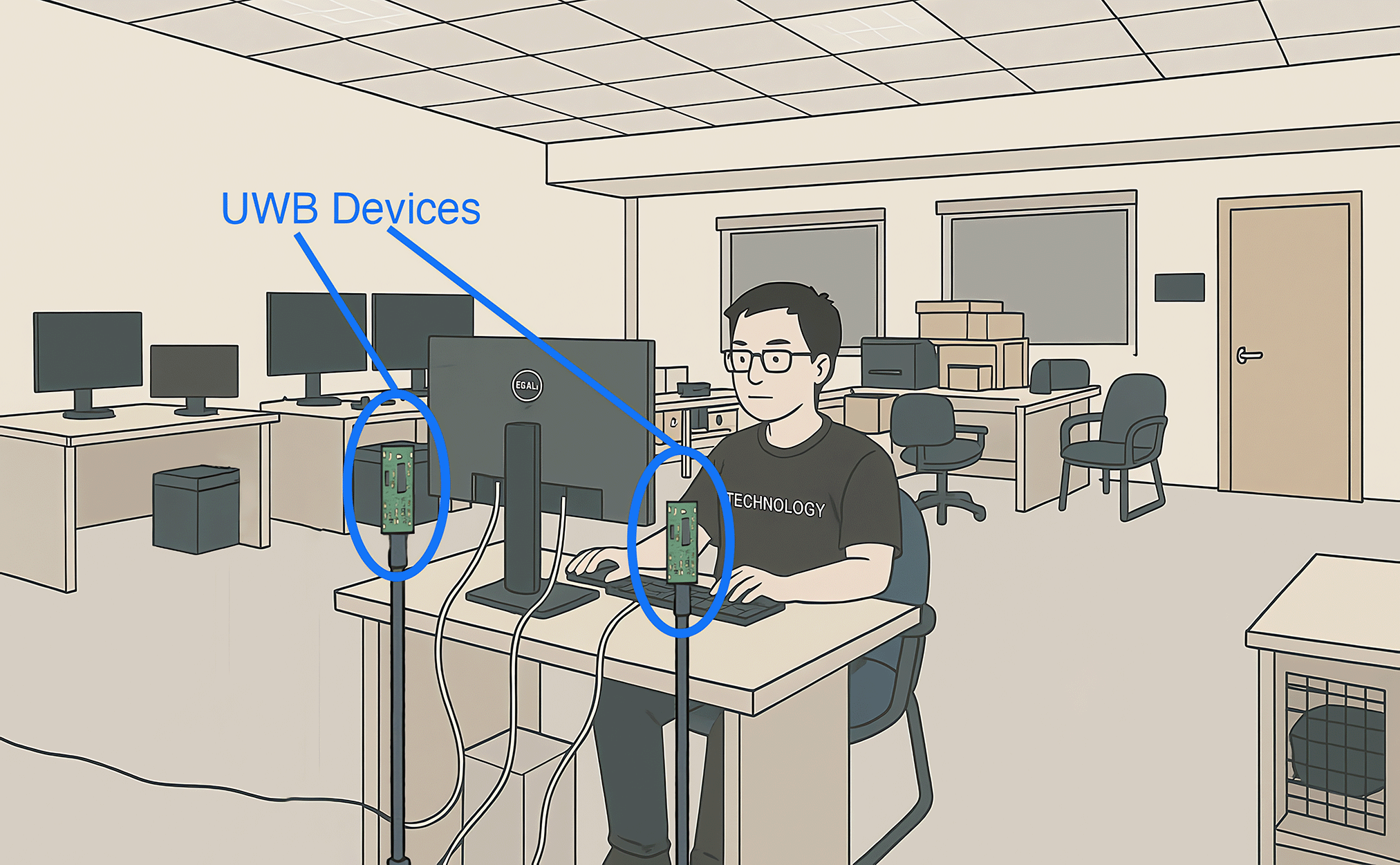}
    \caption{Experiment scenario setup.}
    \label{fig:exp-setup}
\end{figure}

A pair of commercial UWB devices are placed at the edges of the desk, positioned 1 meter apart, flanking the monitor.
The devices are set to a height of 1 meter, with antennas directed at the volunteer.
We further investigate UWB-PostureGuard's robustness to placement variations by adjusting the devices' height, orientation, separation distance, and their distance from the desk edge.
Both devices are connected to a controller for data collection and timestamp synchronization. The monitor is also connected to the controller, delivering instructional prompts to the volunteers.

\subsection{Hardware Configuration}
UWB-PostureGuard is implemented using two Murata Type2BP evaluation boards as UWB devices, a Raspberry Pi 5 as the controller, and another Raspberry Pi 5 with Camera Module 3 to record ground truth. 

The UWB devices are both configured to operate on Channel 9, which has a center frequency of 7987.2 MHz. This channel is selected over other commonly available options, such as Channel 5, due to its superior multi-path resolvability and higher Channel Impulse Response (CIR) resolution.
The slot duration is set to 2 ms, and the ranging interval is set to 200 ms to improve the stability of the data collection process. The preamble code index is set to 10, which works on Base PRF mode, and the preamble duration is set to 64 symbols. We also set one STS segment per ranging frame. The STS segment could secure the synchronization, enable accurate timestamp extraction, and enhance the channel estimation.
For the CIR capture, Post STS CIR data is recorded in both RX1 and RX2.
By featuring high-SNR reflections, Post STS CIR data yields the dual benefits of avoiding preamble interference while providing detailed multi-path information proximal to the main signal path.

\subsection{Dataset Construction}
Data collection for each volunteer follows a standardized protocol. Initially, each participant is provided with the study’s purpose and procedures, after which informed consent is obtained.
Participants are then instructed to assume their natural, seated working posture to establish a baseline. Following this, they are guided through a sequence of predefined postures for recording. To ensure accuracy, two separate baseline recordings, each containing 19 postures described in Section \ref{subsec:posture} without any environmental variations, are captured for each participant. The protocol then requires each posture set with variations to be sequentially recorded once per environmental scenario. 

For a systematic evaluation, participants are divided into two groups: a full recording group, for which all environmental variation scenarios are recorded; and a partial recording group, for which a selected subset of scenarios are recorded.
The per-participant session duration is 90 minutes for the full recording group and 40 minutes for the partial recording group, respectively.
Each recorded sample consists of 100 continuous UWB data frames for a given posture.

We recruit 10 participants (7 male, 3 female; age range: 24 to 34 years, mean: 28), representing a diverse range of body types (height: 168 to 190 cm; weight: 56 to 90 kg; shoulder width: 34 to 48 cm) to ensure realistic dataset variability.
Detailed information about the volunteers is listed in Table \ref{tab:subject_details}, where F means full recording group, and P means partial recording group.
In total, the collected dataset comprises 1,710 samples (171,000 frames), consisting of both labeled and Out-of-Distribution (OOD) data, of which 380 samples constitute the baseline set.

\begin{table}[ht]
\scriptsize
\centering
\caption{Information of the volunteers.}
\label{tab:subject_details}
\begin{tabularx}{\linewidth}{>{\centering\arraybackslash}X 
                              >{\centering\arraybackslash}X 
                              >{\centering\arraybackslash}X 
                              >{\centering\arraybackslash}X 
                              >{\centering\arraybackslash}X 
                              >{\centering\arraybackslash}X}
\toprule
\textbf{Volunteer No. / Group} & \textbf{Age} & \textbf{Height (cm)} & \textbf{Weight (kg)} & \textbf{Shoulder width (cm)} & \textbf{Gender} \\
\midrule
1 / F  & 24 & 176 & 79 & 37 & Male   \\
2 / F  & 34 & 182 & 77 & 38 & Male   \\
3 / F  & 26 & 177 & 90 & 40 & Male   \\
4 / F  & 29 & 190 & 85 & 48 & Male   \\
5 / P  & 26 & 173 & 56 & 40 & Female \\
6 / P  & 24 & 174 & 85 & 43 & Male   \\
7 / P  & 27 & 173 & 82 & 43 & Male   \\
8 / F  & 31 & 168 & 56 & 39 & Female \\
9 / P  & 27 & 177 & 83 & 38 & Female \\
10 / P & 27 & 171 & 58 & 34 & Male   \\
\bottomrule
\end{tabularx}
\end{table}

\subsection{Model Configuration}
\label{sec:exp_method}
For the lightGBM module in our PoseGBDT model, we set the number of leaves to 64, the learning rate to 0.05, and the feature fraction to 0.9. The maximum training step is set to 1000, with an early stopping mechanism when the loss value does not improve in the next 10 steps. We also utilize softmax cross-entropy as the loss function in the training phase:
\begin{align}
    \label{eq:loss}
    L_{\log}(Y, \hat{P}) &= -\log \text{Pr}(Y|\hat{P}) \nonumber \\
    &= - \frac{1}{N} \sum_{i=0}^{N-1} \sum_{k=0}^{K-1} y_{i,k} \log \hat{p}_{i,k}.
\end{align}

Notably, the sliding window size of time series data is set to 5 as discussed in \ref{subsec:models}, which means that each input of the model contains the feature of the current frame, the shifting feature of the past 4 frames, and the rolling mean value of the past 4 frames.

Performance in each experiment is quantified using the same metrics detailed in Section \ref{subsec:models}, namely, overall Accuracy, and the per-posture metrics of Precision, Recall, and F1-Score.
To mitigate potential classification bias in the model, the data is randomly divided into training (60\%) and testing (40\%) sets. Care was taken to preserve the relative proportions of each posture class within both splits.
\section{Evaluation}
\label{sec:evaluation}

\subsection{Model Performance}
\label{sec:overall_perf}
Following the experimental configuration detailed in Section \ref{sec:exp_setup}, we proceed with model training and evaluation.
The overall performance comparison between the standard LightGBM baseline (LG) and PoseGBDT (Ours) is presented in Table \ref{tab:tau_comparison}.
PoseGBDT outperforms the baseline across all posture categories. For static postures, PoseGBDT demonstrates exceptional accuracy, with all metrics above 99\% and minimal inter-class variance.
The performance advantage of PoseGBDT is particularly salient for dynamic postures, which substantially outperforms the baseline, especially for challenging postures (e.g., stretch, and horizontal leg shaking), where the improvements are often by significant margins.
Both macro and weighted averages exceed 99\%, confirming the strong overall performance and robustness of PoseGBDT.

\begin{table}[ht]
\centering
\caption{Performance comparison across methods.}
\label{tab:tau_comparison}
\scriptsize
\begin{tabularx}{\linewidth}{l *{2}{>{\centering\arraybackslash}X} *{2}{>{\centering\arraybackslash}X} *{2}{>{\centering\arraybackslash}X}}
\toprule
\textbf{Posture} & \multicolumn{2}{c}{\textbf{Precision}} & \multicolumn{2}{c}{\textbf{Recall}} & \multicolumn{2}{c}{\textbf{F1-Score}} \\
 & LG & \textbf{Ours} & LG & \textbf{Ours} & LG & \textbf{Ours} \\
\midrule
\multicolumn{7}{c}{static} \\
\midrule
Idle               & 99.99 & 99.99 & 99.99 & 99.99 & 99.99 & 99.99 \\
Upright            & 92.64 & 99.70 & 95.26 & 99.55 & 93.93 & 99.63 \\
Leaning forward    & 95.08 & 99.55 & 96.17 & 99.99 & 95.62 & 99.78 \\
Leaning back       & 86.49 & 99.52 & 90.78 & 99.56 & 88.58 & 99.54 \\
Lateral lean left  & 87.93 & 99.74 & 89.59 & 99.49 & 88.75 & 99.62 \\
Cross leg left     & 84.09 & 99.43 & 86.98 & 99.49 & 85.51 & 99.46 \\
Lateral lean right & 87.89 & 99.60 & 89.59 & 99.75 & 88.73 & 99.67 \\
Cross leg right    & 88.55 & 99.59 & 87.65 & 99.48 & 88.10 & 99.53 \\
Hunch              & 92.28 & 99.70 & 92.19 & 99.80 & 92.24 & 99.75 \\
Tense              & 89.38 & 99.60 & 89.08 & 99.35 & 89.23 & 99.47 \\
Lie on table       & 97.96 & 99.95 & 99.28 & 99.99 & 98.61 & 99.97 \\
\midrule
\multicolumn{7}{c}{dynamic} \\
\midrule
Rotate head           & 59.86 & 97.70 & 63.59 & 98.84 & 61.67 & 98.27 \\
Vert. leg shake left  & 71.88 & 98.70 & 73.58 & 98.75 & 72.72 & 98.72 \\
Vert. leg shake right & 67.84 & 98.72 & 72.51 & 98.12 & 70.10 & 98.42 \\
Horiz. leg shaking    & 57.14 & 97.96 & 58.69 & 99.19 & 57.90 & 98.57 \\
Tap finger            & 88.43 & 99.84 & 89.87 & 98.99 & 89.14 & 99.41 \\
Stretch               & 52.58 & 97.11 & 48.37 & 98.07 & 50.39 & 97.59 \\
Stand                 & 80.57 & 97.85 & 61.80 & 94.83 & 69.95 & 96.31 \\
Walk                  & 88.61 & 98.87 & 59.13 & 97.76 & 70.93 & 98.31 \\
\midrule
Macro Avg.    & 82.59 & 99.11 & 81.27 & 99.00 & 81.69 & 99.05 \\
Weighted Avg. & 81.61 & \textbf{99.11} & 81.58 & \textbf{99.11} & 81.45 & \textbf{99.11} \\
\bottomrule
\end{tabularx}
\end{table}

\begin{figure*}[ht]
    \centering
    \begin{subfigure}[t]{0.36\textwidth}
        \includegraphics[width=\linewidth]{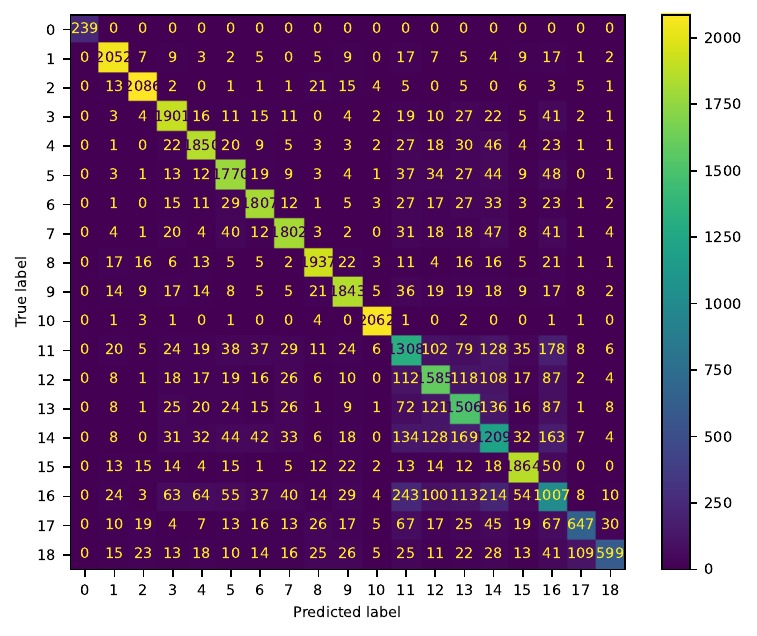}
        \caption{Standard LightGBM Baseline}
    \end{subfigure}
    \hfill
    \begin{subfigure}[t]{0.36\textwidth}
        \includegraphics[width=\linewidth]{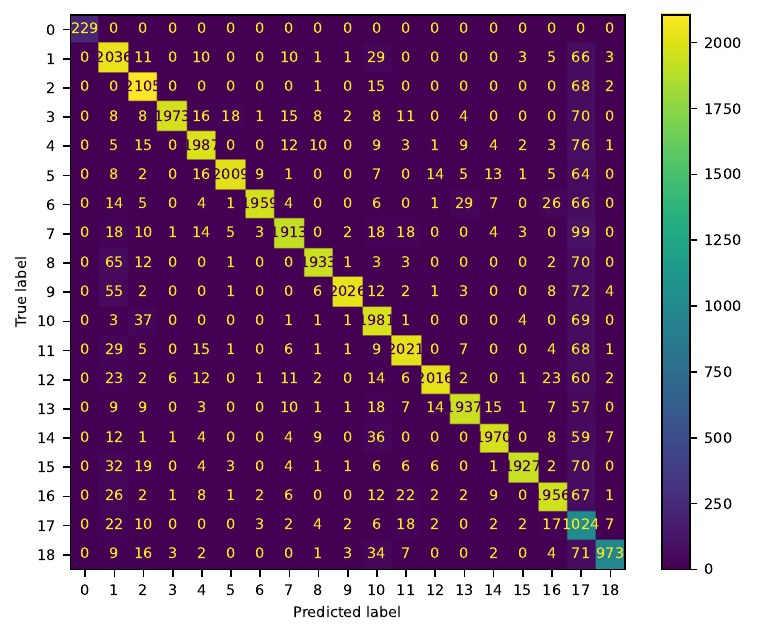}
        \caption{PoseGBDT}
    \end{subfigure}
    \hfill
    \begin{subfigure}[t]{0.27\textwidth}
        \tiny
        \centering
        \begin{tabular}[b]{|c|l|}
            \hline
            0 & Idle \\ \hline
            1 & Upright \\ \hline
            2 & Leaning forward \\ \hline
            3 & Leaning back \\ \hline
            4 & Lateral lean left \\ \hline
            5 & Cross leg left \\ \hline
            6 & Lateral lean right \\ \hline
            7 & Cross leg right \\ \hline
            8 & Hunch \\ \hline
            9 & Tense \\ \hline
            10 & Lie on table \\ \hline
            11 & Rotate head \\ \hline
            12 & Vert. leg shaking left \\ \hline
            13 & Vert. leg shaking right \\ \hline
            14 & Horiz. leg shaking \\ \hline
            15 & Tap finger \\ \hline
            16 & Stretch \\ \hline
            17 & Stand \\ \hline
            18 & Walk \\ \hline
        \end{tabular}
    \end{subfigure}
    \caption{Confusion matrix of standard LightGBM baseline and PoseGBDT.}
    \label{fig:method_comparison}
\end{figure*}

Figure \ref{fig:method_comparison} provides the confusion matrices of each model, which show that PoseGBDT significantly reduces misclassifications compared to the baseline model, particularly in dynamic postures. For example, the baseline model frequently confuses “vertical leg shaking left” with “vertical leg shaking right,” while PoseGBDT distinguishes them accurately.
Overall, PoseGBDT exhibits two key advantages: higher classification accuracy across all postures, and more effective separation of similar posture classes.

\subsection{Impact of Environmental Variations}
\label{sec:ablation}
To explore the impact of different variables on the performance of UWB-PostureGuard, we conduct a series of studies. 
Each studied experimental variable is evaluated using a distinct cohort of five participants.
The default setting in Table \ref{tab:ablation_sv} and \ref{tab:ablation_dv} means PoseGBDT is trained and evaluated on the entire dataset, while the other entries show the performance when evaluating data corresponding to each variable.

\textbf{User Variables} are variables related to real-world user scenarios. We compare several common use cases in Table \ref{tab:ablation_sv}. Specifically, Earbuds refer to each user wearing a pair of Bluetooth earbuds. Smartphone means each user puts a UWB-enabled mobile device on the desk or in their pocket. Smartwatch means that each user wears a UWB-enabled smartwatch on their wrist. Pillow means each user sits holding a pillow in front of their body, and Backpack means each user sits putting a backpack behind their body. Clothing contains two situations, thin (default) and thick, evaluating the penetration capability of UWB signal. Laptop considers a possible use case where each user is using an additional laptop on the desk, which results in an obstacle between UWB sensors and the user. A Water Bottle is another type of obstacle object on the desk. Finally, Chair Type considers two types of chairs, a stack chair (default) and a wheeled chair, both of which have arms.

The results indicate that most user variables exert limited influence on system performance. However, a discernible reduction in accuracy is observed in scenarios involving a laptop or a smartphone. The minor performance decrease associated with the laptop is likely attributable to increased signal blockage, whereas the smartphone's presence may introduce mild signal interference or scattering effects.

\begin{table}[!ht]
    \scriptsize
    \centering
    \caption{Impact of user variable.}
    \label{tab:ablation_sv}
    \begin{tabularx}{\linewidth}{lcccc}
        \toprule
        Variable & Precision & Recall & F1-Score & Accuracy \\
        \midrule
        Default Setting     & 99.11 & 99.11 & 99.11 & 99.11 \\
        \midrule
        Earbuds             & 97.62 & 97.42 & 97.27 & 97.42 \\
        Smartphone (desk)   & 93.46 & 90.14 & 90.92 & 90.14 \\
        Smartphone (pocket) & 94.27 & 90.37 & 91.34 & 90.37 \\
        Smartwatch         & 99.45 & 99.41 & 99.41 & 99.41 \\
        Pillow              & 95.78 & 95.02 & 95.04 & 95.02 \\
        Backpack            & 94.60 & 95.29 & 94.60 & 94.69 \\
        Clothing (thick)    & 94.14 & 90.35 & 91.15 & 90.35 \\
        Laptop              & 93.72 & 85.31 & 87.85 & 85.31 \\
        Water Bottle        & 98.51 & 97.62 & 97.69 & 97.62 \\
        Chair Type (wheeled chair) & 94.33 & 92.26 & 92.67 & 92.26 \\
        \bottomrule
    \end{tabularx}
\end{table}

\textbf{Device Variables} are about the hardware setups of UWB devices. In this part, we consider the distance between UWB devices and desk edges (Dist. Ant-Desk), the default value is 0 cm, which means the projection positions of UWB devices are flush with the two edges of the desk. The horizontal distance between UWB devices (Dist. Ant-Ant) is 100 cm by default. The direction of antennas of both devices (Antenna Orient) changes from face-to-user (by default) to face-to-face. The antenna height is the height of both devices relative to the floor; the default value is 100 cm.

The results for device variables show that changing device deployment leads to modest reductions in all the metrics. Notably, different antenna heights cause the most significant performance drop among these variables, as the main lobe of the antennas is no longer aligned with the user, resulting in some signal energy missing the body and thereby reducing the strength of posture-related features.

\begin{table}[!ht]
    \scriptsize
    \centering
    \caption{Impact of device variable.}
    \label{tab:ablation_dv}
    \begin{tabularx}{\linewidth}{lcccc}
        \toprule
        Variable & Precision & Recall & F1-Score & Accuracy \\
        \midrule
        Default Setting            & 99.11 & 99.11 & 99.11 & 99.11 \\
        \midrule
        Dist. Ant.-Desk (50 cm)    & 94.31 & 93.73 & 93.57 & 93.73 \\
        Dist. Ant.-Ant. (125 cm)   & 98.07 & 97.93 & 97.94 & 97.93 \\
        Ant. Orient (Face-to-Face) & 95.04 & 93.86 & 93.68 & 93.86 \\
        Antenna Height (120 cm)    & 92.61 & 91.35 & 91.67 & 91.35 \\
        \bottomrule
    \end{tabularx}
\end{table}

In summary, the system's performance is largely unaffected by most user and device variables. The most notable exceptions are the presence of a laptop or smartphone, which can cause minor accuracy drops due to signal blockage or interference, and changes in antenna height, which can lead to signal misalignment. Despite these specific factors, the system demonstrates high accuracy and robustness across all tested environmental variations.

\subsection{Continuous Monitoring Demonstration}
\begin{figure}[ht]
    \centering
    \includegraphics[width=\linewidth]{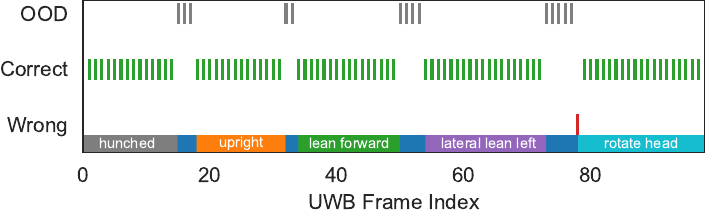}
    \caption{Continuous posture monitoring.}
    \label{fig:case_study}
\end{figure}

In Figure \ref{fig:case_study}, we present a real-world case study that demonstrates the effectiveness of our UWB-PostureGuard system in a continuous monitoring scenario. In this user session, UWB data is collected at 0.2-second intervals as the user transitions sequentially through five distinct postures: hunched, upright, lean forward, lateral lean left, and rotate head. The bottom color bar indicates the true posture labels over time; the dark color represents the unlabeled data.

The results highlight the system’s strong continuous monitoring ability that most frames are classified correctly, even when the user switches postures. Notably, the system is able to promptly detect OOD frames, for example, when a user's behavior does not fit any of the trained classes. This case study demonstrates both the high accuracy and robust OOD detection capability of our system in realistic, continuous scenarios.

\section{Discussion}
\label{sec:discussion}

\textbf{Temporal-Aware Modeling.}
Traditional standard frame-wise classification approaches lack capturing the inherent temporal dependencies in human posture patterns. Our proposed PoseGBDT model addresses this limitation through three key innovations: (1) sliding window strategies that incorporate historical context, (2) time-lagged features that capture posture transitions, and (3) rolling statistical summaries that smooth short-term variations. The significant performance improvement from 81\% (standard LightGBM) to 99.11\% (our PoseGBDT) validates the importance of temporal modeling in posture recognition. The optimal window size of $\tau=5$ frames balances historical context with computational efficiency, preventing overfitting while maintaining discriminative power.

\textbf{Privacy-Preserving Healthcare Monitoring.}
UWB-PostureGuard offers compelling privacy advantages over existing approaches. Unlike camera-based systems that capture identifiable visual information, our RF-based approach processes only electromagnetic signal reflections that cannot be reverse-engineered to reveal personal identity or appearance. This privacy-by-design implementation addresses critical workplace surveillance concerns while enabling continuous health monitoring. The contactless nature eliminates the discomfort and compliance issues associated with wearable sensors, making it suitable for long-term deployment in sensitive environments such as healthcare facilities and corporate offices.

\textbf{Real-World Deployment Feasibility.}
The system demonstrates practical viability with a total cost under \$50 using commercially available UWB modules measuring only 3cm × 3cm. This compact form factor enables seamless integration into existing office environments without requiring infrastructure modifications. Our extensive environmental variation studies reveal that most variables (e.g., thick clothing, large obstacles) have limited impact on the performance, the system maintains over 90\% accuracy, and can be highly adaptable across diverse real-world scenarios.

\textbf{Limitations and Future Directions.}
Despite achieving high accuracy, our system faces several limitations that warrant future investigation. First, the current implementation focuses on single-user scenarios and may experience interference in multi-user environments. Second, while the system effectively monitors torso and lower body postures, limb detection remains limited due to the placement strategy and the number of sensors. 
In addition, several promising directions emerge for extending this work. Multi-sensor fusion approaches could enable simultaneous monitoring of multiple users while maintaining privacy preservation. Integration with smartphone accelerometers and gyroscopes could provide complementary limb posture information. Also, long-term behavioral analysis using our temporal modeling framework could identify gradual posture degradation patterns and provide personalized intervention strategies. 

\section{Conclusion}
\label{sec:conclusion}
In this paper, we develop UWB-PostureGuard, a privacy-preserving UWB sensing system for continuous sitting posture monitoring, which includes the PoseGBDT model to achieve robust, high-accuracy classification across 19 postures without using cameras or wearable devices. Experimental evaluations show that our system maintains strong performance and resilience under various real-world conditions, while the system’s compact and affordable hardware design supports practical deployment. This work demonstrates the potential of RF-based sensing for health monitoring, and future efforts will focus on further extending system versatility and granularity.



\printbibliography

\end{document}